\documentclass[10pt,twocolumn,letterpaper]{article}

\usepackage{cvpr}
\usepackage{times}
\usepackage{epsfig}
\usepackage{graphicx}
\usepackage{amsmath}
\usepackage{booktabs}
\usepackage{amssymb}
\usepackage{multirow}
\usepackage{pifont}
\usepackage{adjustbox}

\newcommand{{\xmark}}{{\ding{55}}}
\newcommand{{\cmark}}{{\ding{51}}}%

\usepackage[pagebackref=true,breaklinks=true,letterpaper=true,colorlinks,bookmarks=false]{hyperref}

\cvprfinalcopy 

\def\cvprPaperID{7959} 

\begin{document}

\title{\textsc{Violin}: A Large-Scale Dataset for Video-and-Language Inference}

\author{Jingzhou Liu$^{1}$$^{*}$ \quad Wenhu Chen$^{2}$$^{*}$  \quad Yu Cheng$^{3}$ \quad Zhe Gan$^{3}$   \quad Licheng Yu$^{3}$ \\
Yiming Yang$^{1}$   \quad Jingjing Liu$^{3}$\\
    $^{1}$Carnegie Mellon University  \quad $^{2}$University of California, Santa Barbara \\
    $^{3}$Microsoft Dynamics 365 AI Research \\
    {\tt\small \{liujingzhou,yiming\}@cs.cmu.edu, wenhuchen@ucsb.edu}\\
    {\tt\small \{yu.cheng,zhe.gan,licheng.yu,jingjl\}@microsoft.com }\\
    }

\maketitle

\begin{abstract}
We introduce a new task, Video-and-Language Inference, for joint multimodal understanding of video and text. Given a video clip with aligned subtitles as premise, paired with a natural language hypothesis based on the video content, a model needs to infer whether the hypothesis is entailed or contradicted by the given video clip. A new large-scale dataset, named \textsc{Violin} (VIdeO-and-Language INference), is introduced for this task, which consists of 95,322 video-hypothesis pairs from 15,887 video clips, spanning over 582 hours of video. These video clips contain rich content with diverse temporal dynamics, event shifts, and people interactions, collected from two sources: ($i$) popular TV shows, and ($ii$) movie clips from YouTube channels. In order to address our new multimodal inference task, a model is required to possess sophisticated reasoning skills, from surface-level grounding (e.g., identifying objects and characters in the video) to in-depth commonsense reasoning (e.g., inferring causal relations of events in the video). We present a detailed analysis of the dataset and an extensive evaluation over many strong baselines, providing valuable insights on the challenges of this new task. 
\end{abstract}
{\let\thefootnote\relax\footnote{{$^{*}$This work was done while the authors were interns at Microsoft.}}}

\begin{figure*}
\begin{center}
\includegraphics[width=\linewidth]{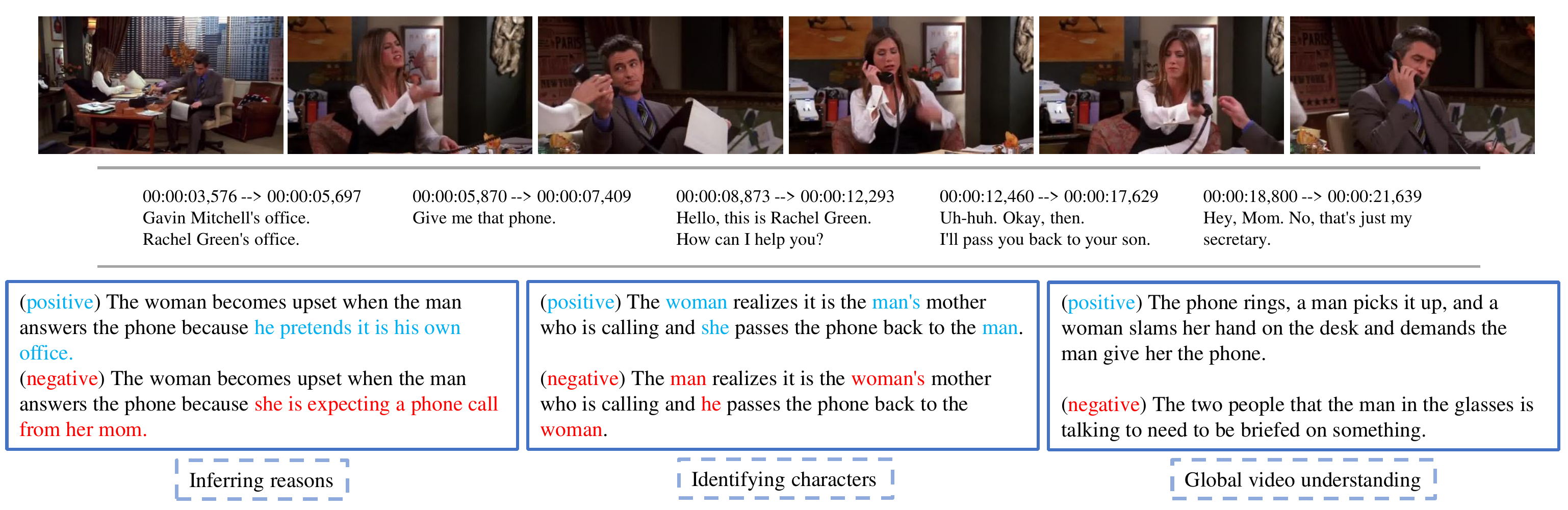}
\end{center}
\vspace{-3mm}
   \caption{An example from the \textsc{Violin} dataset. The first two rows show a video clip with its aligned subtitles. The third row contains three pairs of positive/negative statements. The task is to independently decide whether each statement is supported or contradicted given the subtitled video. The first two negative statements are written by modifying part of the positive statements (marked in red), and the third is curated by adversarial matching (Sec.~\ref{sec: dataset}). The text box below each pair of statements indicates the reasoning skill required to infer the verdict of each statement.}
\label{fig:example}
\vspace{-3mm}
\end{figure*}
\section{Introduction}
%
Joint vision-and-language understanding sits at the nexus of computer vision and natural language processing (NLP), and has attracted rapidly growing attention from both communities. Popular tasks include visual question answering~\cite{antol2015vqa,goyal2017making}, referring expression comprehension~\cite{yu2016modeling,yu2018mattnet}, visual dialog~\cite{das2017visual}, visual reasoning~\cite{johnson2017clevr,suhr2018corpus,hudson2019gqa}, visual commonsense reasoning~\cite{zellers2019recognition}, NLVR$^2$~\cite{suhr2018corpus}, and visual entailment~\cite{xie2019visual}. The emergence of these diverse Vision+Language tasks, benchmarked over large-scale human annotated datasets~\cite{lin2014microsoft,krishna2017visual}, has driven tremendous progress in joint multimodal embedding learning~\cite{tan2019lxmert,lu2019vilbert,chen2019uniter,su2019vl}. 
However, most of these datasets and models were centered on static images, 
leaving the joint modeling of video and its aligned textual information (\emph{e.g.}, video-and-language understanding) a relatively under-explored territory. 

Video Question Answering (Video QA) is one of the most popular tasks in current studies for video-and-language understanding.
Video QA model aims to answer a natural language question given a video clip. 
Existing Video QA datasets include MovieFIB~\cite{maharaj2017dataset}, MovieQA~\cite{tapaswi2016movieqa}, TGIF-QA~\cite{jang2017tgif}, PororoQA~\cite{kim2017deepstory}, and TVQA~\cite{lei2018tvqa,lei2019tvqaplus}. 
While these datasets have covered a rich pool of video content (\emph{e.g.}, cartoons, short GIFs and TV shows), they are limited to QA task only.
On the other hand, in NLP field, one important benchmark for natural language understanding is natural language inference (NLI)~\cite{bowman2015large,williams2017broad}, where a model is presented with a pair of sentences (premise and hypothesis), and judges the relationship between the pair (\emph{e.g.}, \emph{Contradiction}, \emph{Neutral}, and \emph{Entailment}).

Inspired by NLI,
we present a novel task, Video-and-Language Inference, to foster deeper investigations in video-and-language understanding. Specifically, given a video clip with aligned subtitles as premise, and a natural language statement as a hypothesis describing the video content, a model is expected to infer whether the statement is entailed or contradicted by the given video clip.
This new task is easy to evaluate, since only binary classification is measured; but also challenging to solve, as a thorough interpretation of both visual and textual clues is required in order to achieve in-depth understanding and inference for a complex video scenario. 

We introduce a large-scale dataset for this new task, \textbf{VI}de\textbf{O}-and-\textbf{L}anguage \textbf{IN}ference (\textsc{Violin})\footnote{Project page: \url{https://github.com/jimmy646/violin}.}, built upon natural video content with rich temporal dynamics and social interactions. Video clips are collected from diverse sources to cover realistic visual scenes, and statements are collected from crowdsource workers via Amazon Mechanical Turk (AMT)\footnote{\url{https://www.mturk.com/}}, who watched the videos accompanied by subtitles (dialogue, scene description, etc). Our goal is to provide a dataset that can test a model's cross-modality reasoning skills over both video and textual signals. To this end, we require AMT workers to write statements based on joint understanding of both video and subtitles, which not only describe explicit information in the video (\emph{e.g.}, objects, locations, characters, social activity), but also reveal in-depth comprehension of complex plots (\emph{e.g.}, interpreting human emotions and relations, understanding the events, inferring causal relations of events throughout the video). This distinguishes our collected statements from the straightforward captions in video/image captioning dataset~\cite{lin2014microsoft,krishna2017dense,wang2019vatex}, which are dominated by explicit factual descriptions without deeper inference.

Writing negative statements for an inference task is challenging~\cite{bowman2015large,zellers2019recognition}. To gather high-quality negative statements without artificial cues or biased priors, we employed two strategies in the data collection: ($i$) requiring annotators to write negative statements by changing just a few words or phrases in a positive statement, to ensure that the style and length of the statement remain unchanged; ($ii$) performing adversarial matching~\cite{zellers2019recognition}: for each video, select challenging and confusing statements from the statement pool of other videos as the negative ones. The first strategy ensures the collected statements can test a model's in-depth inference ability, since only a small fraction of a positive statement is modified, which requires the model to distinguish highly similar statements with different meanings. The second strategy focuses more on testing a model's global understanding of the video, to distinguish statements with high-level scene difference between videos. When combined together, these two strategies produce a dataset with minimal visual or textual bias. Through this effort, we collected 95,322 video-statement pairs, containing 15,887 video clips spanning over 582 hours. Each video is paired with 6 statements and is 35.2 seconds long on average.

The main contributions of this paper are three-fold. ($i$) We propose a new task, Video-and-Language Inference, which requires a model to draw inference on whether a written statement entails or contradicts a given video clip. ($ii$) We introduce a new dataset \textsc{Violin} for this task, providing a reliable benchmark for measuring joint video-and-language understanding models. 
($iii$) We provide a detailed analysis of the \textsc{Violin} dataset with evaluation over strong baselines, and suggest future directions for this new task.

\begin{table*}
\begin{center}
\small
\scalebox{0.9}{
\begin{tabular}{l c c c c c c }
\hline
source & \# episodes & \# clips & avg clip len & avg pos. statement len & avg neg. statement len & avg subtitle len\\
\hline
Friends & 234 & 2,676 & 32.89s & 17.94 & 17.85 & 72.80\\
Desperate Housewives & 180 & 3,466 & 32.56s & 17.79 & 17.81 & 69.19\\
How I Met Your Mother & 207 & 1,944 & 31.64s & 18.08 & 18.06 & 76.78\\
Modern Family & 210 & 1,917 & 32.04s & 18.52 & 18.20 & 98.50\\
MovieClips & 5,885 & 5,885 & 40.00s & 17.79 & 17.81 & 69.20\\
\hline
All & 6,716 & 15,887 & 35.20s & 18.10 & 18.04 & 76.40\\
\hline
\end{tabular}}
\end{center}
\vspace{-2mm}
\caption{Statistics of different video sources used to create our dataset.}
\label{tab:data-stat}
\vspace{-2mm}
\end{table*}

\section{Related Work}
\paragraph{Natural Language Inference (NLI)}
Understanding entailment and contradiction relations between sentences (\emph{i.e.}, Natural Language Inference) is fundamental to natural language understanding. Several large-scale datasets have been developed as NLI benchmarks, such as SNLI~\cite{bowman2015large} and MultiNLI~\cite{williams2017broad}. NLI is also included in the GLUE benchmark for evaluating general language understanding~\cite{wang2018glue}. Recent introduction of large-scale pre-trained language models, such as BERT~\cite{devlin2018bert}, XLNet~\cite{yang2019xlnet}, and RoBERTa~\cite{liu2019roberta}, has propelled significant progress in NLI. Multi-task learning and adversarial training~\cite{liu2019multi,zhu2019freelb} also prove to be helpful in improving model performance. 

Inspired by NLI, we propose the task of Video-and-Language Inference to evaluate a system's multimodal reasoning ability. However, different from NLI, our task is more challenging in the sense that both video and text (subtitles) are provided; thus, a thorough joint understanding of both modalities is required for inference. 

\vspace{5pt}
\noindent\textbf{Visual Entailment} \,
Visual Entailment (VE)~\cite{xie2019visual} is a recently proposed task that extends NLI to the visual domain. In this task, a natural image premise and a natural language hypothesis are given, and the goal is to judge whether the textual hypothesis can be confirmed based on the visual content in the image. Three labels are assigned: \emph{Entailment, Neutral}, and \emph{Contradiction}. The dataset is created based on Flickr30k image captions~\cite{young2014image} and SNLI~\cite{bowman2015large}. Similarly, NLVR$^2$~\cite{suhr2018corpus} is proposed to investigate the grounding relationship between given images and a natural language description.

Our proposed task is different from VE in the following aspects. ($i$) VE considers images as input, while our task focuses on videos instead. Compared with static images, videos contain complex temporal dynamics, making the video-and-language inference task  more challenging as the model needs to understand the relationship between different visual scenes to draw inference. ($ii$) Our proposed task requires deeper visual understanding. Images in the VE task are mostly natural images, while the videos in \textsc{Violin} were collected from popular TV shows and movie clips, which contain rich social interactions and diverse scenes. This requires a model to not only understand explicit visual cues, but also infer in-depth rationale behind the scene. ($iii$) Our task requires more sophisticated language understanding. VE is a combination of Flickr30k~\cite{young2014image} and SNLI~\cite{bowman2015large}, with no crowdsouring involved. The hypotheses in VE task are composed of captions only, containing factual descriptions that can be explicitly derived from the visual content in the image. On the other hand, \textsc{Violin} mainly consists of implicit statements that cannot be solved without in-depth understanding of the video and text, designed specifically to evaluate a model's multimodal reasoning skills. 

\vspace{5pt}
\noindent\textbf{Video-and-Language Research} \,
With the emergence of large-scale video datasets~\cite{caba2015activitynet,abu2016youtube,kay2017kinetics,videvent,Wang_2016_CVPR}, several video-and-language tasks have been proposed, such as video captioning~\cite{guadarrama2013youtube2text,venugopalan2015sequence,xu2016msr,gan2017semantic,krishna2017dense,gan2017stylenet,pu2018adaptive,wang2019vatex}, localizing video segments from natural language queries~\cite{gao2017tall,anne2017localizing,chen2018temporally,lei2020tvr}, video reasoning~\cite{yi2019clevrer}, and video question answering \cite{tapaswi2016movieqa,lei2018tvqa}. Video captioning is a conditional text generation task, while the other three belong to video-and-language understanding. In particular, MovieQA~\cite{tapaswi2016movieqa}, TGIF-QA~\cite{jang2017tgif} and TVQA~\cite{lei2018tvqa,lei2019tvqaplus}, which contain real-world videos and human-generated questions, are recently proposed for video question answering. 

Our \textsc{Violin} dataset also uses TV shows as one of the video sources, similar to TVQA~\cite{lei2018tvqa}. The main differences are summarized as: ($i$) Our dataset contains richer video content, including 5,885 movie clips in additional to TV shows used in TVQA. 
($ii$) Our dataset requires more sophisticated reasoning skills from a model, such as inferring reasons and interpreting human emotions, while most QA pairs in TVQA are focused on identifying explicit information.

\vspace{5pt}
\noindent\textbf{Visual Question Answering} \,
Our proposed task is also related to Visual Question Answering (VQA)~\cite{antol2015vqa,goyal2017making}. The CLEVR dataset~\cite{johnson2017clevr} serves as a popular synthetic diagnosis dataset that tests a model's compositional reasoning skills. Recently, GQA~\cite{hudson2019gqa} was introduced to benchmark real-world visual reasoning, and VCR~\cite{zellers2019recognition} for visual commonsense reasoning. 

Many neural network models have been proposed for these tasks, such as more advanced attention mechanisms~\cite{yang2016stacked,lu2016hierarchical,yu2019deep}, better multimodal fusion methods~\cite{fukui2016multimodal,yu2017multi,kim2016hadamard,kim2018bilinear}, the use of multi-step reasoning~\cite{hudson2018compositional,gan2019multi,cadene2019murel}, the incorporation of relations~\cite{santoro2017simple,li2019relation,norcliffe2018learning}, and neural module networks for compositional reasoning~\cite{andreas2016neural,johnson2017inferring,hu2017learning,chen2019meta}.  
Our proposed task can provide a new perspective for benchmarking these models. 

\section{Video-and-Language Inference Dataset}

\begin{table*}
\begin{center}
\small
\scalebox{0.9}{
\begin{tabular}{l c c c c c c c }
\hline
Dataset & Visual Domain & Source & Subtitles & Inference & Task & \# images/videos & \# samples\\
\hline
Movie-QA~\cite{tapaswi2016movieqa} & video & movie & \cmark & \xmark & QA & 6.8K & 6.5K\\
MovieFIB~\cite{maharaj2017dataset} & video & movie& \xmark & \xmark & QA & 118.5K & 349K\\
TVQA~\cite{lei2018tvqa} & video & TV show & \cmark & \xmark & QA & 21.8K & 152.5K\\
VCR~\cite{zellers2019recognition} & image & movie & \xmark & \cmark & QA & 110K & 290K\\
GQA~\cite{hudson2019gqa} & image & indoor & \xmark & \cmark & QA & 113K & 22M \\
SNLI-VE~\cite{xie2019visual} & image & natural & \xmark & \cmark & Entailment & 31.8K & 565.3K\\
NLVR$^2$~\cite{suhr2018corpus} & image& natural & \xmark & \cmark & Entailment & 127.5K & 107.3K\\
\hline
\textsc{Violin} (ours) & video & TV show/movie & \cmark & \cmark & Entailment & 15.9K & 95.3K\\
\hline
\end{tabular}}
\end{center}
\vspace{-2mm}
\caption{Comparison between \textsc{Violin} and other existing vision-and-language datasets.}
\label{tab:data-comp}
\vspace{-2mm}
\end{table*}
In our \textsc{Violin} dataset for video-and-language inference, the input is a video clip $V$ consisting of a sequence of video frames $\{v_i\}_{i=1}^T$, paired with its aligned text $S=\{s_i, t_i^{(0)},t_i^{(1)}\}^n_{i=1}$ ($s_i$ is the subtitle within time span $(t_i^{(0)}\rightarrow t_i^{(1)})$ in the video) and a natural language statement $H$ as the hypothesis aiming to describe the video clip. For every $(V,S,H)$ triplet, a system needs to perform binary classification: $f(V,S,H)\rightarrow\{0,1\}$, deciding whether the statement $H$ is entailed (label $1$) from or contradicts (label $0$) the given video clip. In order to increase the coverage and versatility, we collect the videos from diverse sources, including 4 popular TV shows of different genres and YouTube movie clips from thousands of movies. To ensure high video quality, we also provide carefully-designed protocols to guide crowdsource workers to select representative video segments for which to write positive/negative statements. The procedure of dataset collection is detailed in Sec.~\ref{sec: dataset}, and Sec.~\ref{sec: analysis} provides a thorough analysis on the dataset.

\subsection{Dataset Collection} \label{sec: dataset}
\begin{figure}[t!]
\begin{center}
\hspace{0cm}  
   \includegraphics[width=0.85\linewidth]{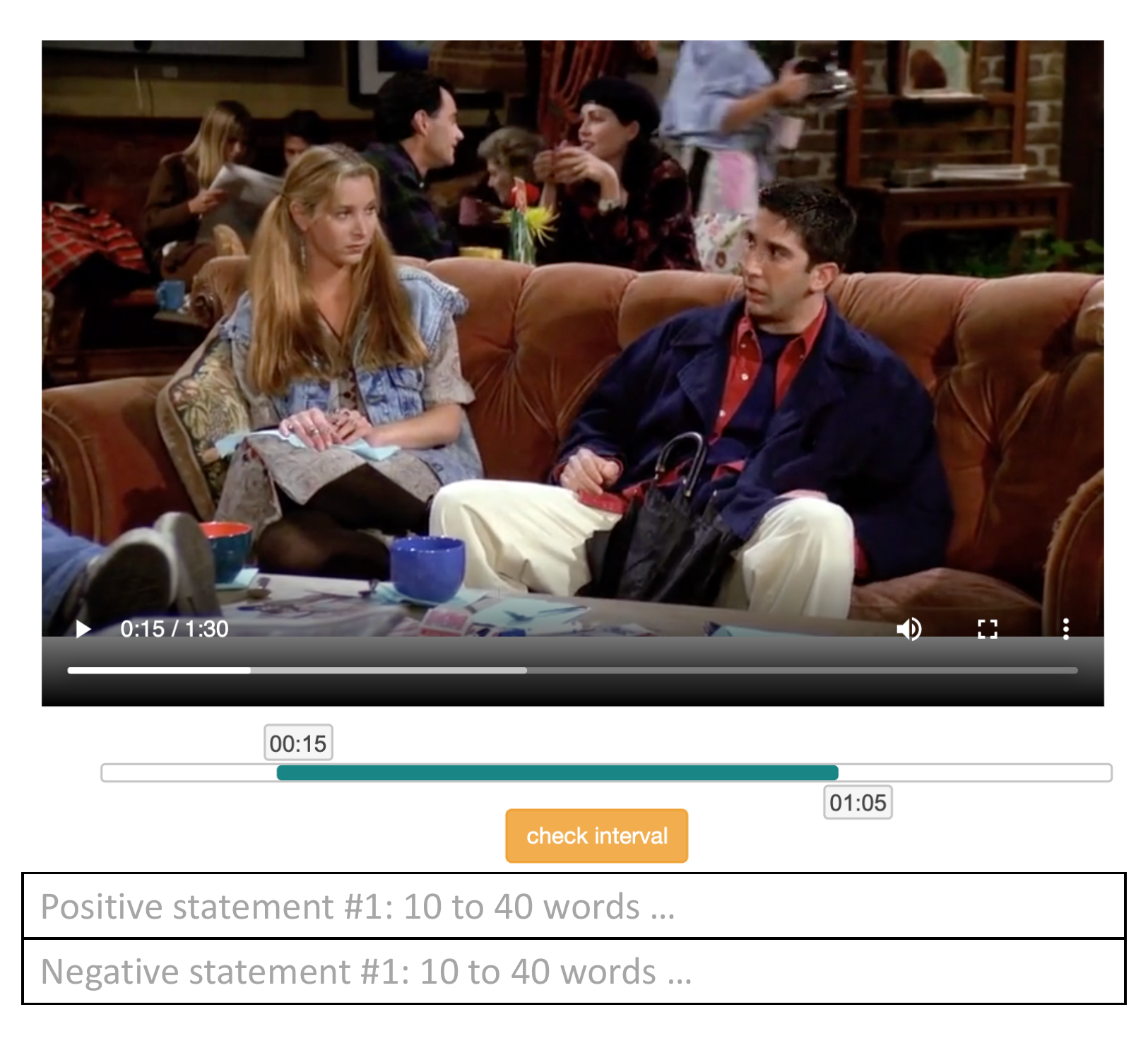}
\end{center}
\vspace{-6mm}
\caption{User interface for annotators. Each annotator is provided with a video clip and required to first drag the slider below the video player to select a single-scene clip from the video, then write three pairs of positive/negative statements in the text boxes}
\label{fig:amt-ui}
\vspace{-3mm}
\end{figure}

We collect videos from two sources: ($i$) 4 popular TV shows, 
and ($ii$) movie clips from YouTube channels\footnote{\url{https://www.youtube.com/user/movieclips}} covering thousands of movies. Both sources contain rich human interactions and activities. Each episode of the TV shows is 20-40 minutes long, which we split into clips of 90 seconds long (while avoiding splitting dialogues in the middle). These 90 second-long clips may contain more than one scene, which are then presented to crowdworkers to select a video segment containing a single, self-contained scene for which they can write the statements. Additionally, we restrict the length of the selected interval to 15-40 seconds long, to maintain a reasonable difficulty level for the task. For movie clips from YouTube channels, the original lengths are around two minutes, which by nature usually contain only one scene of the movie. Thus, there is no need for the workers to manually select a video segment from the provided movie clips. We just select the first 40 seconds from every movie clip for annotation, to keep it consistent with TV show clips. Figure~\ref{fig:amt-ui} shows the interface for AMT workers. By dragging the slider below the video player, users can adjust the start and end timestamps of the segment they want to select (for movie clips the slider is disabled).

After video segments are selected, they are presented to another group of annotators to write positive/negative statements. Each worker is assigned with one video clip, and is required to write three pairs of positive/negative statements describing the video (in the text boxes in Figure~\ref{fig:amt-ui}). We do not require AMT workers to follow any templates, as our goal is to collect diversified and natural expressions. We do have several rules/guidelines for writing positive statements: ($i$) We do not allow annotators to refer to characters in the video by name. Instead, they should use grounded referring expressions (\emph{e.g.}, ``the man with blonde hair wearing grey shirt", ``the girl sitting in the sofa holding a cup of coffee"). The purpose of this is to keep the dataset consistent across different video sources (not all video clips have character names), and to reduce potential bias (in TV shows, the number of character names is very small). ($ii$) We ask workers to keep to a minimum level of copying from subtitles (\emph{e.g.}, ``somebody says ...") or describing explicit visual information (\emph{e.g.}, object, color), and encourage them to write statements combining information from both the video clip and subtitles. ($iii$) We encourage workers to write about different aspects of the given video clip in different statement pairs, which may require different types of reasoning, such as inferring character emotions/relations/intentions and inferring causal relations in complex events.


In practice, we observe that when letting human annotators write negative statements without any constraint, the resulting statements show serious bias (\emph{i.e.}, models can learn to classify positive/negative statements without even absorbing information from the video or subtitles). When intentionally writing fake content without any reference, humans tend to use subtle patterns that statistical models can easily pick up. Therefore, when collecting negative statements, we propose two strategies to alleviate the bias issue. First, we ask annotators to use a positive statement as reference, and only modify a small portion of it to make it negative. In this case, most part of the statement remains true to the video content, and human-introduced bias is kept to minimum. This rigorous setting makes the statements more challenging to distinguish by the model, and in-depth reasoning is required to identify the fake content. 
For quality control, only workers located in English-speaking countries with a lifetime task approval rate greater than 98\% can participate in our study. Also, during data collection, we manually check every worker's submissions to ensure the quality of the video segments and statements.

VCR~\cite{zellers2019recognition} proposes adversarial matching to construct wrong answers for multiple-choice QA, by selecting a correct answer (from another question) that is most similar to the current question. In our task, we use a similar strategy. For a human-generated positive statement $H_i$ for video $V_i$, we select a positive statement $H_j$ collected for another video $V_j$, which is most similar to $H_i$, and use $(H_i, H_j)$ as a pair of positive/negative statements for video $V_i$. Using this strategy, a portion of the collected statements serve as both positive and negative samples, which helps removing artificial bias. Unlike the first strategy aforementioned, statement pairs constructed this way focus more on the global understanding of the video. For example, in Figure~\ref{fig:example}, the first two negative statements are written by modifying positive statements (the modified part is marked in red), and the third negative statement is obtained by adversarial matching. In the final dataset, $2/3$ of the negative statements are constructed following the first strategy, and the remaining $1/3$ with the second strategy.

\subsection{Dataset Analysis} 
\label{sec: analysis}

The \textsc{Violin} dataset contains 15,887 video clips, and each video clip is annotated with 3 pairs of positive/negative statements, resulting in 95,322 $(V,S,H)$ triplets in total. Statistics on the full dataset is provided in Table~\ref{tab:data-stat}. Each statement has 18 words on average, and the lengths of positive and negative statements are almost the same, showing no significant bias in length.

\begin{figure}[t!]
\begin{center}
\hspace{-0.5cm}  
   \includegraphics[width=\linewidth]{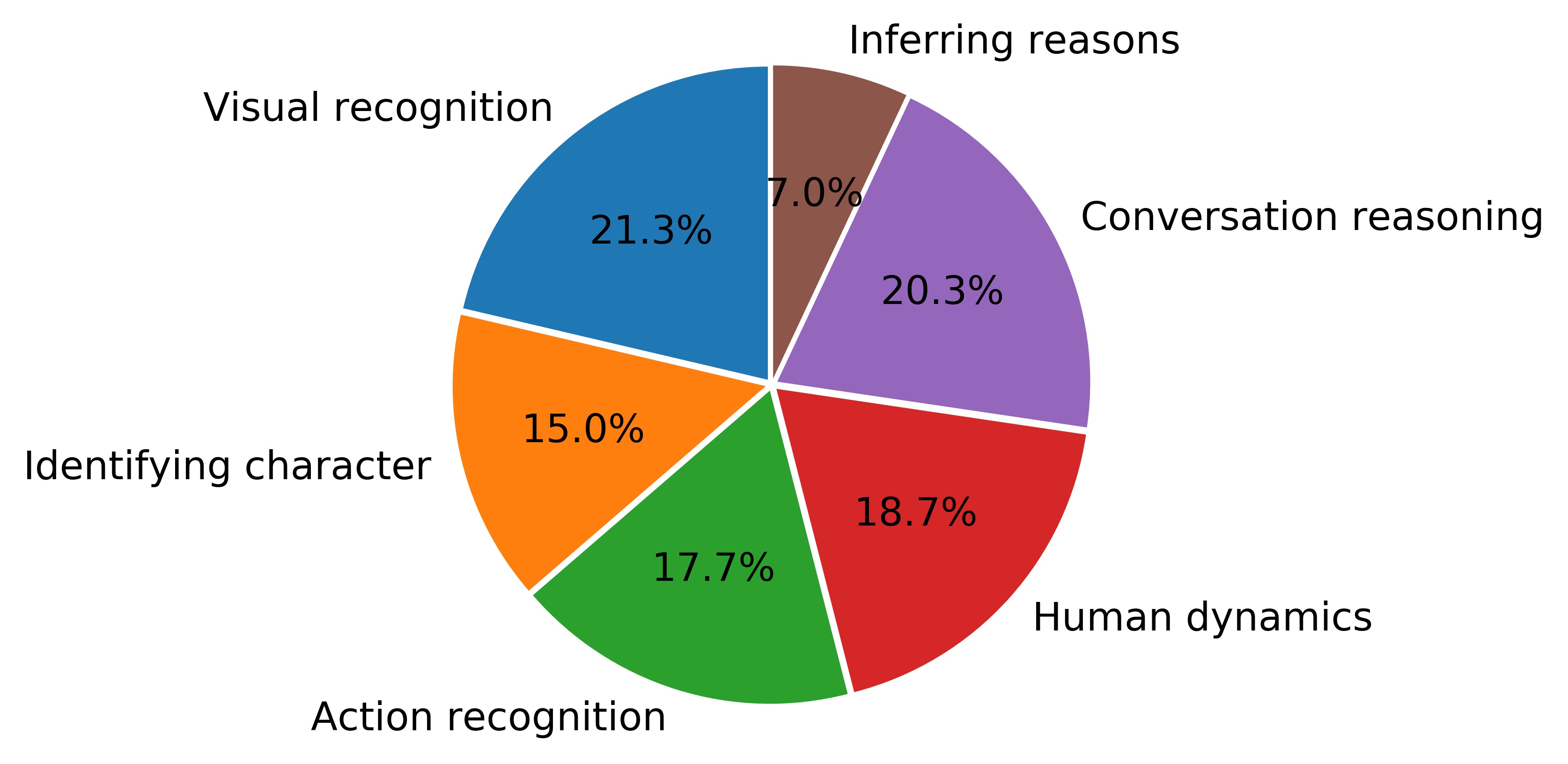}
\end{center}
\vspace{-3mm}
   \caption{Distribution of reasoning types. ``Visual recognition", ``identifying character" and ``action recognition" focus on explicit visual information; the other three require high-level inference.}
\label{fig:category}
\end{figure}

As discussed in Sec.~\ref{sec: dataset}, we use two strategies to collect negative statements: one is adversarial matching that tests a model's ability of global video understanding; the other is modifying a small part of a positive statement for the video clip, which requires in-depth reasoning skills for a model to distinguish between positive and negative statements. To investigate in more detail, for each pair of positive and negative statements, we categorize it into 6 types of reasoning skills required, as shown in Figure~\ref{fig:category}. The types of ``visual recognition", ``identifying character", and ``action recognition" are more focused on explicit information and require relatively low-level reasoning. ``Human dynamics" includes inferring human emotions/relations/intentions, etc. ``Conversation reasoning" requires performing inference over characters' dialogues and other forms of interactions (body language, hand gestures, etc.). And ``inferring reasons" is about inferring causal relations in complex events. These 3 types of statement require in-depth understanding and commonsense reasoning. Overall, ``explicit information recognition" makes up 54\% of the dataset, and ``commonsense reasoning" makes up the remaining 46\%, making our dataset a balanced one, imposing new challenges on multi-facet video-and-language understanding. Compared to other datasets, our \textsc{Violin} dataset is more focused on reasoning rather than surface-level grounding (\emph{e.g.}, in TVQA~\cite{lei2018tvqa}, only 8.5\% of the questions require reasoning).

\section{Model}
\label{sec:model}

\begin{figure*}[t!]
\begin{center}
\includegraphics[width=0.9\linewidth]{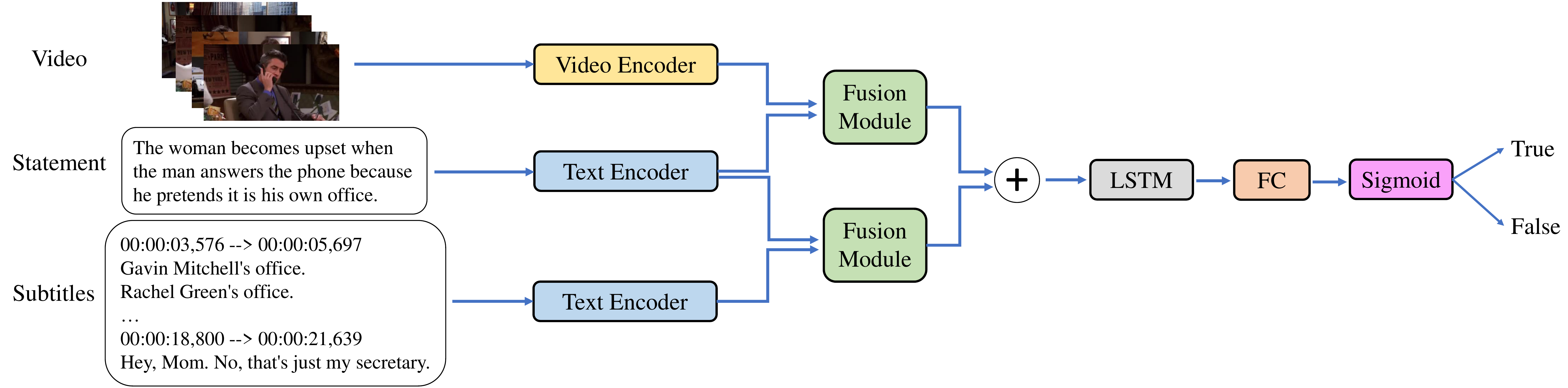}
\end{center}
\vspace{-3mm}
   \caption{Overview of the proposed model for the Video-and-Language Inference task. The model takes a video (a sequence of frames), its aligned subtitles and a statement hypothesis as input, and produces a scalar measuring the probability of the input statement being positive.}
\label{fig:model}
\end{figure*}

In this section, we introduce our baseline model used for benchmarking the \textsc{Violin} dataset and evaluating the effectiveness of different feature choices. An overview of the model is illustrated in Figure~\ref{fig:model}.

\subsection{Video and Text Encoders}
We first extract a sequence of visual features from video frames as $\mathbf{V}\in\mathbb{R}^{T\times d_v}$, where $T$ is the number of time steps, and $d_v$ is the dimension of each feature. Choices of visual features will later be discussed in Sec.~\ref{sec:exp-baselines}. The video encoder is implemented by a bi-directional LSTM, to capture the temporal correlation among consecutive frames. By passing video features into the video encoder and stacking hidden states from both directions, we obtain the video representations as $\mathbf{H}_V\in\mathbb{R}^{T\times2d}$, where $d$ is the hidden-state dimension of the LSTM encoder.

Statements and subtitles share the same text encoder. Statements are tokenized into a word sequence $\{w_i\}_{i=1}^{n_{stmt}}$. Each line in the subtitle is tokenized, and all the lines are concatenated together into one single word sequence $\{u_i\}_{i=1}^{n_{subtt}}$. Here, $n_{stmt}$ and $n_{subtt}$ are the lengths of statement and subtitle, respectively. We experiment with two types of text encoder: LSTM encoder and BERT~\cite{devlin2018bert} encoder. For LSTM encoder, every word token is converted to its word embedding and then fed to the LSTM encoder, producing text representations $\mathbf{H}_{stmt}\in\mathbb{R}^{n_{stmt}\times2d}$ and $\mathbf{H}_{subtt}\in\mathbb{R}^{n_{subtt}\times2d}$. For BERT encoder, we use pre-trained BERT-base model, finetuned on \textsc{Violin} training statements and subtitles. The output of BERT encoder at each position is 768-dimensional, which is then projected to $2d$ dimensions, also denoted as $\mathbf{H}_{stmt}$ and $\mathbf{H}_{subtt}$.

\subsection{Combining Multimodality Streams}
The model takes three streams of information as input: video, subtitles and statement. The goal is to determine whether the statement entails or contradicts with the video and subtitles. In our model, statement representations are jointly modeled with video and subtitles via a shared fusion module. The fusion module is implemented with bidirectional attention, adopted from \cite{seo2016bidirectional,yu2018qanet,lei2018tvqa}, where it is used for query-context matching. For simplicity, we only describe the process of combining the video and the statement streams. Subtitles and statement are fused in a similar way. Statement representations $\mathbf{H}_{stmt}\in\mathbb{R}^{n_{stmt}\times2d}$ are used as context, and video representations $\mathbf{H}_V\in\mathbb{R}^{T\times2d}$ as query. Each word in the statement thus attends to every time step in the video representations. Let $\mathbf{a}_i\in\mathbb{R}^T$ be attention weights for the $i$-th word in the statement, $\sum_{j=1}^T\mathbf{a}_{i,j}=1$ for all $i=1,\ldots,n_{stmt}$, $\mathbf{a}\in\mathbb{R}^{n_{stmt}\times T}$. The output is a video-aware statement representation: $\mathbf{M}_{stmt}^V=\mathbf{a}\mathbf{H}_V\in\mathbb{R}^{n_{stmt}\times2d}$. Similarly, we combine subtitles and statement streams to obtain a subtitle-aware statement representation $\mathbf{M}_{stmt}^{subtt}\in\mathbb{R}^{n_{stmt}\times2d}$. These two sets of representations are further fused via: 
\begin{equation}\nonumber
\resizebox{.48 \textwidth}{!}{
$
\mathbf{M}_{stmt}^{all}=[\mathbf{H}_{stmt};\mathbf{M}_{stmt}^V;\mathbf{M}_{stmt}^{subtt};\mathbf{H}_{stmt}\odot \mathbf{M}_{stmt}^V;\mathbf{H}_{stmt}\odot \mathbf{M}_{stmt}^{subtt}],
$
}
\end{equation}
where $\odot$ stands for element-wise product. The resulting matrix $\mathbf{M}_{stmt}^{all}\in\mathbb{R}^{n_{stmt}\times10d}$ combines information from all three modality streams, which is then fed into another bidirectional LSTM. The last hidden states from both directions are concatenated and passed through a fully-connected layer with 1-dimensional output followed by a sigmoid activation function, predicting the probability of the input statement being positive.

The proposed baseline model is similar to the one in \cite{lei2018tvqa}. The main difference is that our model uses statement representations as context and video/subtitle representations as query in the fusion module. The intuition is that, in our video-and-language inference task, the full statement needs to be supported by evidence from either the video or subtitles, in order to judge the statement to be positive/negative, instead of just locating the position in the video/subtitles that is most relevant to the query (as in TVQA~\cite{lei2018tvqa}). Thus, in our model, every word in the statement is attended to the video and subtitles in the fusion module, then combined and fed to the final bi-LSTM to make the prediction.

\section{Experiments}
\label{sec:exp}
For evaluation, we compare our model with several baselines on the dataset and provide detailed analysis on the results. 
In all the experiments, we split the \textsc{Violin} dataset into 80\% for training (76,122 $(V,S,H)$ triplets), 10\% for validation (9,600 triplets) and 10\% for testing (9,600 triplets). Model performance is evaluated via binary classification accuracy.

\subsection{Compared Models}
\label{sec:exp-baselines}
First, we define the following combinations of input sources, to evaluate the importance of different modality streams:
\vspace{3px}\\
\textbf{Statements Only}: Using statements only, without absorbing information from video or subtitles. This option is to test the innate bias of positive/negative statements.\vspace{3px}\\
\textbf{Video}: Using video features only.\vspace{3px}\\
\textbf{Subtitles}: Using subtitles only.\vspace{3px}\\
\textbf{Video+Subtitles}: Using both video and subtitle features, which is the full setting for the task.\vspace{3px}\\
\textbf{Single Frame+Subtitles}: Using subtitle features plus only one middle frame from the video. This option is to test the usefulness of temporal information in the video.

Different visual features are also evaluated on the \textsc{Violin} task: ($i$) Image feature: we use ResNet101 \cite{He2015} trained on ImageNet \cite{imagenet} to extract the global image feature for each frame;
($ii$) C3D feature: we use 3-dimensional convolutional neural network (C3D) \cite{c3d} to extract video features;
($iii$) Detection feature: we run Faster R-CNN \cite{fasterrcnn} trained on Visual Genome~\cite{krishna2017visual} to detect objects in each frame and use their regional features as the input. 
For image features, we first down-sample each video to 3 frames per second, and then extract the 2048-dim feature for each frame. 
Similarly, for detection features, we use the same sampling rate and extract features followed by a pooling layer outputting the 2048-dim feature for each frame.
For C3D features, we extract 4096-dim features for every 16 frames on the original video (without down-sampling). 
To encode text input as features, we use ($i$) pre-trained BERT-base model~\cite{devlin2018bert} finetuned on \textsc{Violin} statements and subtitles in the training set, and ($ii$) GloVe \cite{glove} embeddings. For thorough evaluation, we also test a large-scale pre-trained model LXMERT~\cite{tan2019lxmert} that jointly learns multimodal features.

\subsection{Experimental Results}
Table \ref{tab:acc} summarizes results from baseline methods and our proposed model (using full-length video clips, subtitles and statements). We also run a set of experiments with different visual/text features and compare the results in Table \ref{tab:acc}. 

\begin{table}[t!]
\begin{center}
\small
\begin{tabular}{c|c|c|c|c}
\hline 
\# & Method & Visual & Text & Accuracy \\
\hline
0 & Random & - & - & 50.00 \\
\hline
1 & Stmt & - & GloVe & 53.94\\
2 & Stmt & - & BERT & \textbf{54.20}\\
\hline
3 & Stmt+Subtt & - & GloVe & 60.10 \\
4 & Stmt+Subtt & - & BERT & \textbf{66.05} \\
\hline
5 & Stmt+Vis & Img & GloVe & 55.30\\
6 & Stmt+Vis & Img & BERT & 59.26\\
7 & Stmt+Vis & C3D & GloVe & 55.91 \\
8 & Stmt+Vis & C3D & BERT & 58.34\\
9 & Stmt+Vis & Det & GloVe & 56.15 \\
10 & Stmt+Vis & Det & BERT & \textbf{59.45} \\
\hline
11 & Stmt+Subtt+SglFrm & Img & BERT & 66.60 \\
\hline
12 & Stmt+Subtt+Vis & Img & GloVe & 60.33\\
13 & Stmt+Subtt+Vis & Img & BERT & 67.60\\
14 & Stmt+Subtt+Vis & C3D & GloVe & 60.68\\
15 & Stmt+Subtt+Vis & C3D & BERT & 67.23\\
16 & Stmt+Subtt+Vis & Det & GloVe & 61.31 \\
17 & Stmt+Subtt+Vis & Det & BERT & \textbf{67.84}\\
\hline
18 & Stmt+Subtt+Vis &\multicolumn{2}{c|}{LXMERT} & 66.25 \\
\hline
\end{tabular}
\end{center}
\vspace{-2mm}
\caption{Accuracy of different methods on \textsc{Violin} test set. Subtt = Subtitle, Vis = Video, Stmt = Statement, SglFrm = single frame,
Img = Image features, Det = Detection features, C3D = C3D features, BERT = BERT features, LXMERT = LXMERT features.}
\label{tab:acc}
\end{table}

\vspace{5pt}
\noindent\textbf{Baseline Comparison}\, Row 0 is the random guess baseline with an accuracy of 50\%. When using only the statement to decide whether itself is positive or negative, the best model with BERT features only achieves 54.20, presenting little bias in the dataset. By adding subtitles or video, 
all the models obtain significant gains over the ``statement only" versions. Notably, Stmt+Subtt with BERT and Stmt+Vis with Det+BERT achieve 66.05 (row 4) and 59.45 (row 10), respectively.
From row 3-4 and 12-17, we can observe that adding subtitles improves the performance significantly. However, the gain of adding video (row 5-10 and 12-17) is not as significant as adding subtitles. This might be due to visual features not capturing video information well. Using only one frame as video features (row 11) is worse than using full video (row 13), showing the importance of exploiting temporal information in the video. Overall, the best performance is achieved by using all the sources, with BERT and Detection features (row 17).

\vspace{5pt}
\noindent\textbf{Model Variants} \, We first evaluate the effectiveness of different visual features. In most settings, Detection features work better than Image and C3D features, indicating that the extracted regional information and external knowledge from Visual Genome are useful for this task.
Among all the textual features, BERT~\cite{devlin2018bert} is the strongest as expected. In all the settings, BERT-based versions generally improve the accuracy by 3\% to 6\%, compared with non-contextualized embedding such as GloVe~\cite{glove}. Joint multimodal embedding (LXMERT, row 18) achieves 66.25, which is slightly worse than the best baseline model (row 17), showing that \textsc{Violin} imposes more challenges on existing single-image-based joint pre-trained models. 

\vspace{5pt}
\noindent\textbf{Human Evaluation} \, Human performance via AMT is presented in Table \ref{tab:human}. As expected, humans achieve the best performance when provided with both video and subtitles (85.20)\footnote{We repeated the human evaluation ourselves, and the accuracy is 93\%.}. Without context (video and subtitles), humans only achieve 51.38\% accuracy. Interestingly, we find that adding video brings in more gain than adding subtitles, showing the importance of visual information in \textsc{Violin} task.   

\begin{table}[t]
\begin{center}
\small
\begin{tabular}{c|c}
\hline 
Source & Test Accuracy (\%)  \\
\hline
Statement & 51.38 \\
Subtitle + Statement & 73.85\\
Video + Statement & 77.19\\
Video+Subtitle+Statement & {85.20} \\
\hline
\end{tabular}
\end{center}
\vspace{-2mm}
\caption{Accuracy in human evaluation on test set over
different input sources.}
\label{tab:human}
\end{table}

\begin{table}[t]
\begin{center}
\small
\begin{tabular}{c|c|c}
\hline 
Method & Annotated & Adversarial matching \\
\hline
Stmt+Subtt & 61.05 & 66.05\\
Stmt+Vis & 57.08 & 59.26 \\
Stmt+Subtt+Vis & 61.99 & 67.60\\
\hline
\end{tabular}
\end{center}
\vspace{-2mm}
\caption{Accuracy (\%) on test set containing negative statements collected via different strategies. Image and BERT features are used in this experiment.}
\label{tab:neg-sacc}
\end{table}

\begin{figure*}
\begin{center}
\includegraphics[width=\linewidth]{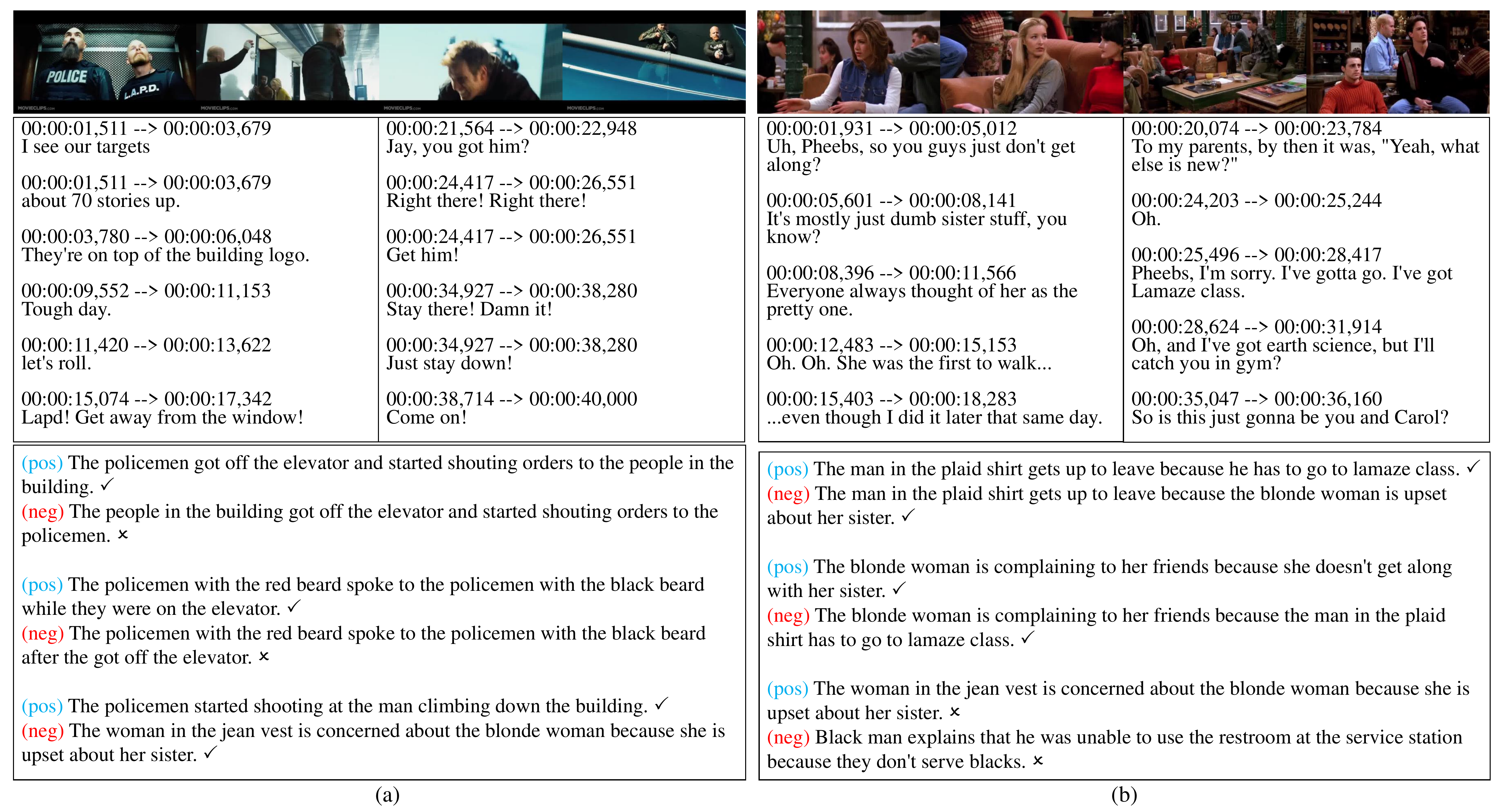}
\end{center}
\vspace{-3mm}
\caption{Qualitative analysis results. Pos/neg at the beginning of each statement indicates ground truth. {\cmark} or {\xmark} at the end of each statement represents model prediction. {\cmark} means the system judges the statement as positive, and {\xmark} means negative.}
\label{fig:qualitative}
\end{figure*}

\begin{table}[t!]
\begin{center}
\small
\begin{adjustbox}{scale=0.88,center}
\begin{tabular}{c|c|c|c|c|c}
\hline
Statement & Stmt+ &\multicolumn{2}{c|}{Stmt+Vis} & \multicolumn{2}{c}{Stmt+Subtt+Vis}\\
\cline{3-6} 
Reasoning Type & Subtt & Img & Det & Img & Det \\
\hline \hline
Visual recognition & 67.19 & 67.41 & 67.41 & 67.97 & 67.97 \\
\hline
Identify character & 57.78 & 64.44 & 65.18 & 62.22 & 62.22 \\
\hline
Action recognition & 70.75 & 66.04 & 66.04 & 73.58 & 73.58 \\
\hline
Human dynamics & 63.39 & 58.04 & 58.04 & 60.71 & 61.48 \\
\hline
Conversation reasoning & 76.23 & 58.20 & 58.20 & 76.23 & 76.23 \\
\hline
Inferring reasons & 59.52 & 50.00 & 50.31 & 59.52 & 60.18 \\
\hline
\end{tabular}
\end{adjustbox}
\end{center}
\vspace{-2mm}
\caption{Accuracy (\%) on each statement type in \textsc{Violin} test set. All the methods use BERT feature. }\label{table:compared_scores}
\end{table}

\subsection{Further Analysis}

\noindent\textbf{Accuracy on Different Question Types} \, To have a better understanding of the dataset, we examine the accuracy of models on different statement types on test set in Table \ref{table:compared_scores}. Compared to Stmt+Subtt, Stmt+Subtt+Vis models improve mostly on ``visual recognition" and ``action recognition". For categories such as ``inferring reasons" and ``identify character",  including video gains some improvement. On ``conversation reasoning" and ``human dynamics", adding video features does not help.

\vspace{5pt}
\noindent\textbf{Human-Written vs. Adversarially-Sampled Negatives} \, For comparison, we create a new statement set by replacing the adversarially-sampled negative statements with original human-written negative statements. Results are presented in Table \ref{tab:neg-sacc}. Performance on the sampled negatives is higher than that on human-written ones. Our interpretation is that human-written content has higher propensity for intent understanding and in-depth reasoning, which makes the statements more challenging to the model. 

\vspace{5pt}
\noindent\textbf{Qualitative Analysis} \, Figure \ref{fig:qualitative} presents some prediction examples from our model using statement, video and subtitles. The correct cases in Figure \ref{fig:qualitative} (a) demonstrate the model's ability to recognize action, infer emotion, identify referred person, and understand temporal dynamics in the video. In (b), the error cases show that our model does not work well on inferring reasons and human relations.   

\section{Conclusion}
We introduce a new task, video-and-language inference (\textsc{Violin}), which requires intelligent systems to capture rich temporal signals about activities/events in video and text, in order to acquire reasoning skills for multimodal inference. We provide thorough baseline experiments for benchmarking different models on the large-scale dataset, as well as a comprehensive analysis of the dataset. 
The gap between the baseline models and human performance is significant. We encourage the community to participate in this task and invent stronger methods to push the state of the art on multimodal inference. Possible future directions include developing models to localize key frames, as well as better utilizing the alignment between video and subtitles to improve reasoning ability.

\vspace{5pt}
\noindent\textbf{Acknowledgement} \,
We would like to thank Yandong Li, Liqun Chen, Shuyang Dai, Linjie Li, Chen Zhu, Jiacheng Xu and Boyi Li for providing useful feedback on the project and their help in collecting and annotating data. We thank all the reviewers for their helpful comments. The first author is supported in part by NSF under grant IIS-1546329.

\clearpage
{\small
\bibliographystyle{ieee_fullname}
\bibliography{reference}
}

\clearpage

\appendix
\section{Additional Data Analysis}
\subsection{Statement Length Distribution}
\begin{figure}[h]
\begin{center}
   \includegraphics[width=\linewidth]{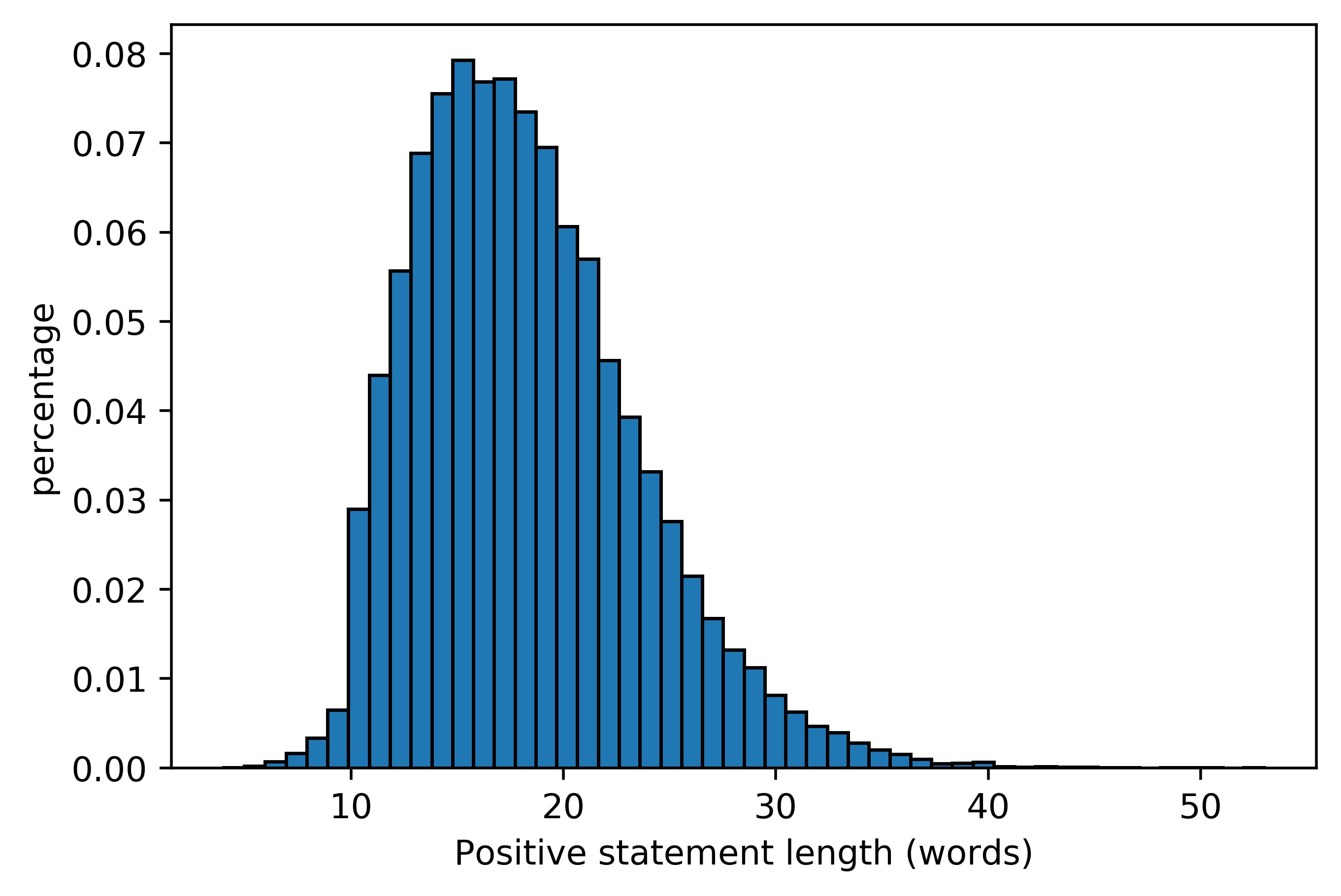}
\end{center}
   \caption{Distribution of positive statement lengths.}
\label{fig:real_len}
\end{figure}

\begin{figure}[h]
\begin{center}
   \includegraphics[width=\linewidth]{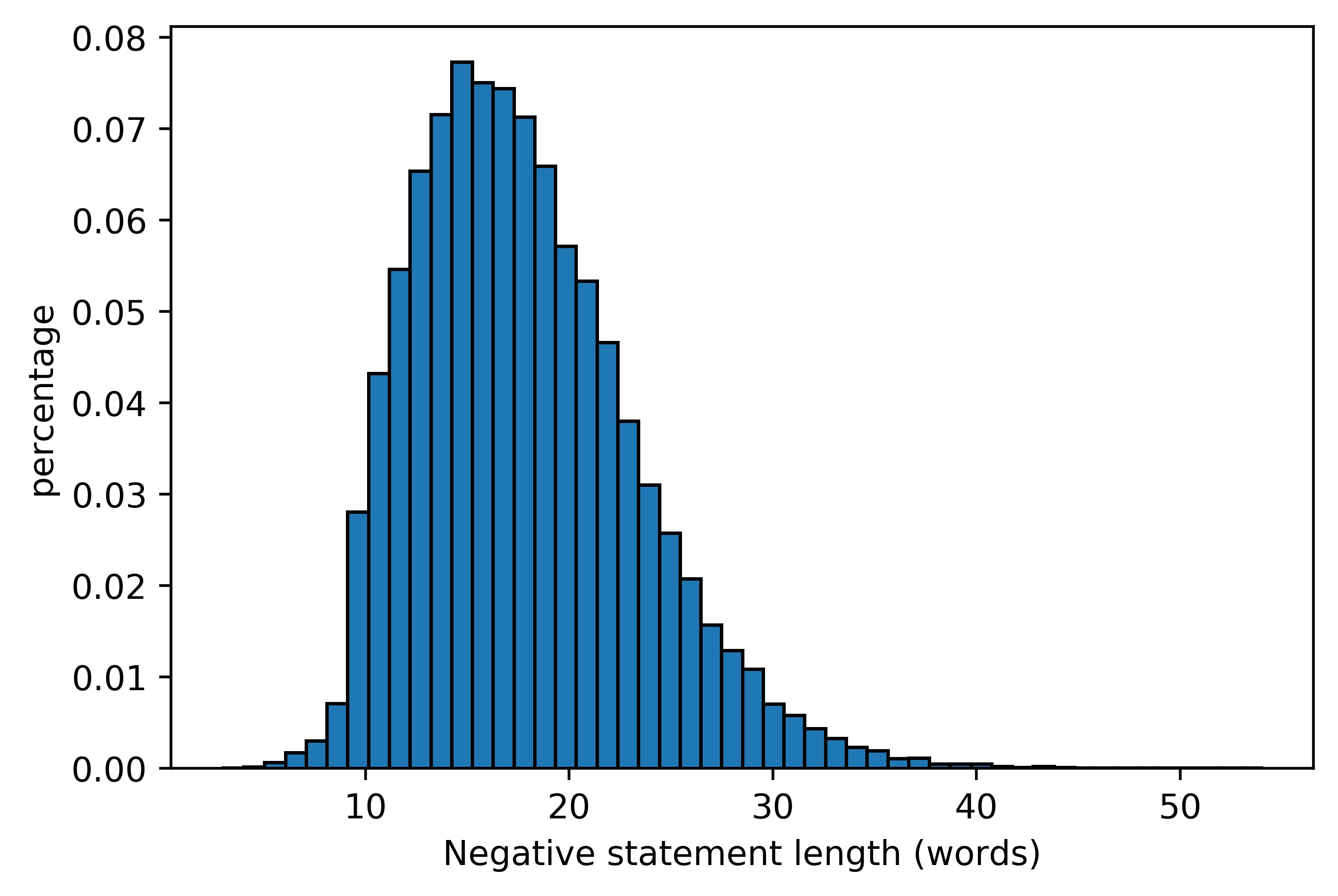}
\end{center}
   \caption{Distribution of negative statement lengths.}
\label{fig:fake_len}
\end{figure}

The length distribution for positive and negative statements are presented in Figure~\ref{fig:real_len} and Figure~\ref{fig:fake_len}, respectively. There is no significant bias in statement lengths for positive and negative statements.
\subsection{Statement Content}
Table~\ref{table:common_words} shows the most common nouns, verbs and adjectives in positive statements, respectively.
\begin{table}[t!]
\begin{center}
\small
\begin{adjustbox}{scale=0.9,center}
\begin{tabular}{c|c}
\hline
Type & Most Common Words \\
\hline
     & man, woman, shirt, suit, hair, jacket, girl, lady, boy, dress,\\
Noun & sweater, friend, brunette, room, guy, people, tie, glass, table,\\
     & car, coat, door, hat, phone, hand, top, bed, house, couch, group\\
\hline
     & tell, wear, ask, want, sit, try, say, talk, go, explain,\\
Verb & walk, get, make, look, see, think, take, give, will, hold,\\
     & can, stand, know, come, leave, feel, have, find, put, like\\
\hline
     & black, blue, blonde, red, white, brown, green, haired, young, dark,\\
Adj  & grey, old, other, pink, purple, upset, plaid, gray, yellow, long, little\\
     & blond, happy, good, excited, surprised, striped, light, angry, short\\
\hline
\end{tabular}
\end{adjustbox}
\end{center}
\caption{Most common words in positive statements.}
\label{table:common_words}
\end{table}
\subsection{Video Length Distribution}
\begin{figure}[h]
\begin{center}
\hspace{-0.5cm}  
   \includegraphics[width=\linewidth]{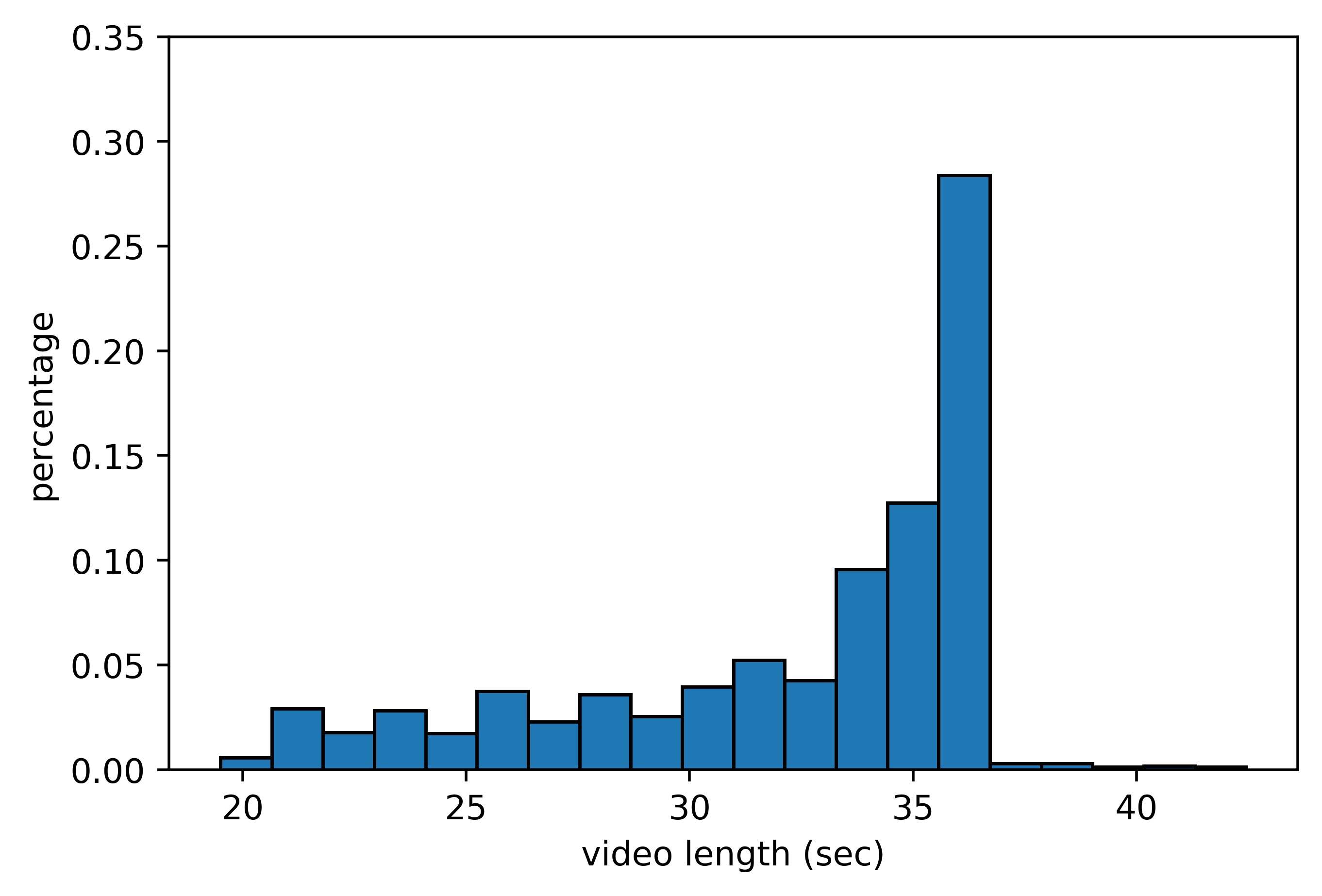}
\end{center}
   \caption{Distribution of video lengths from TV shows.}
\label{fig:vid_len}
\end{figure}

The video clips collected from MovieClips are all 40 seconds long. For video clips collected from TV shows, their lengths vary from 15 to 40 seconds, shown in Figure~\ref{fig:vid_len}.

\section{Instructions for Human Annotators}
\begin{figure*}[h]
\begin{center}
\includegraphics[width=\linewidth]{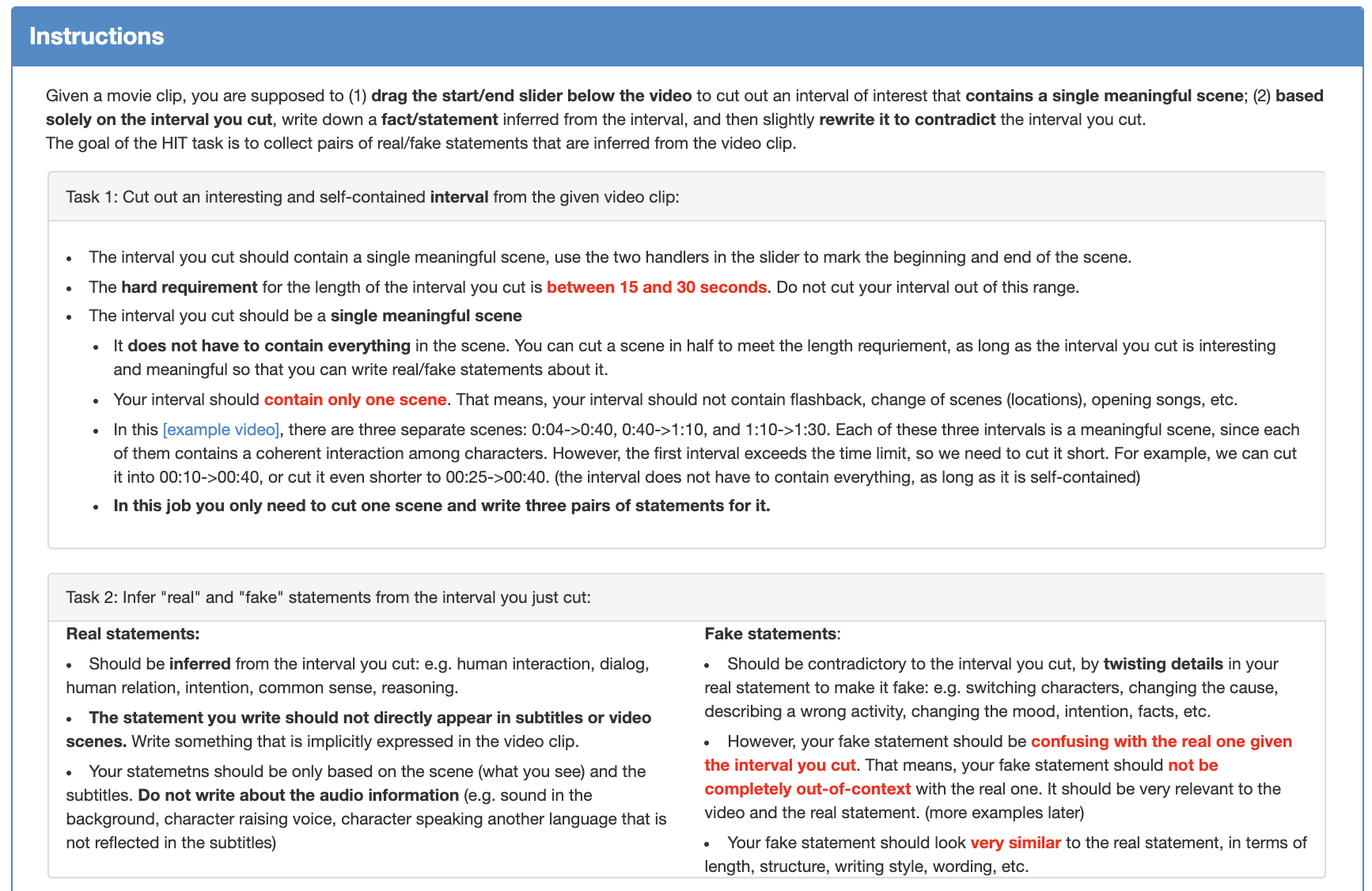}
\end{center}
   \caption{Overall instructions for human annotators.}
\label{fig:ins1}
\end{figure*}

\begin{figure*}[h]
\begin{center}
\includegraphics[width=\linewidth]{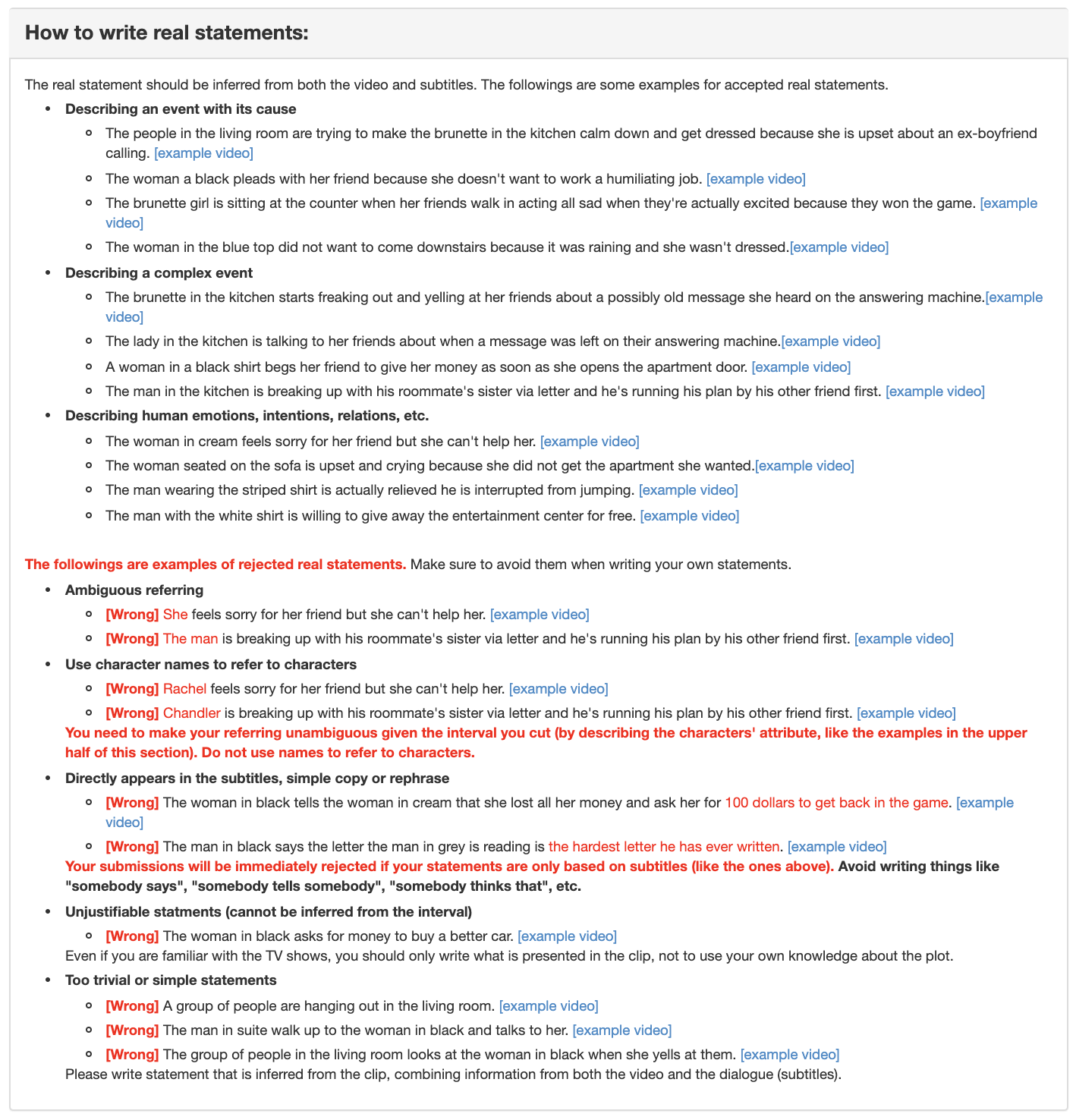}
\end{center}
   \caption{Instructions for writing real statements.}
\label{fig:ins2}
\end{figure*}

\begin{figure*}[h]
\begin{center}
\includegraphics[width=0.9\linewidth]{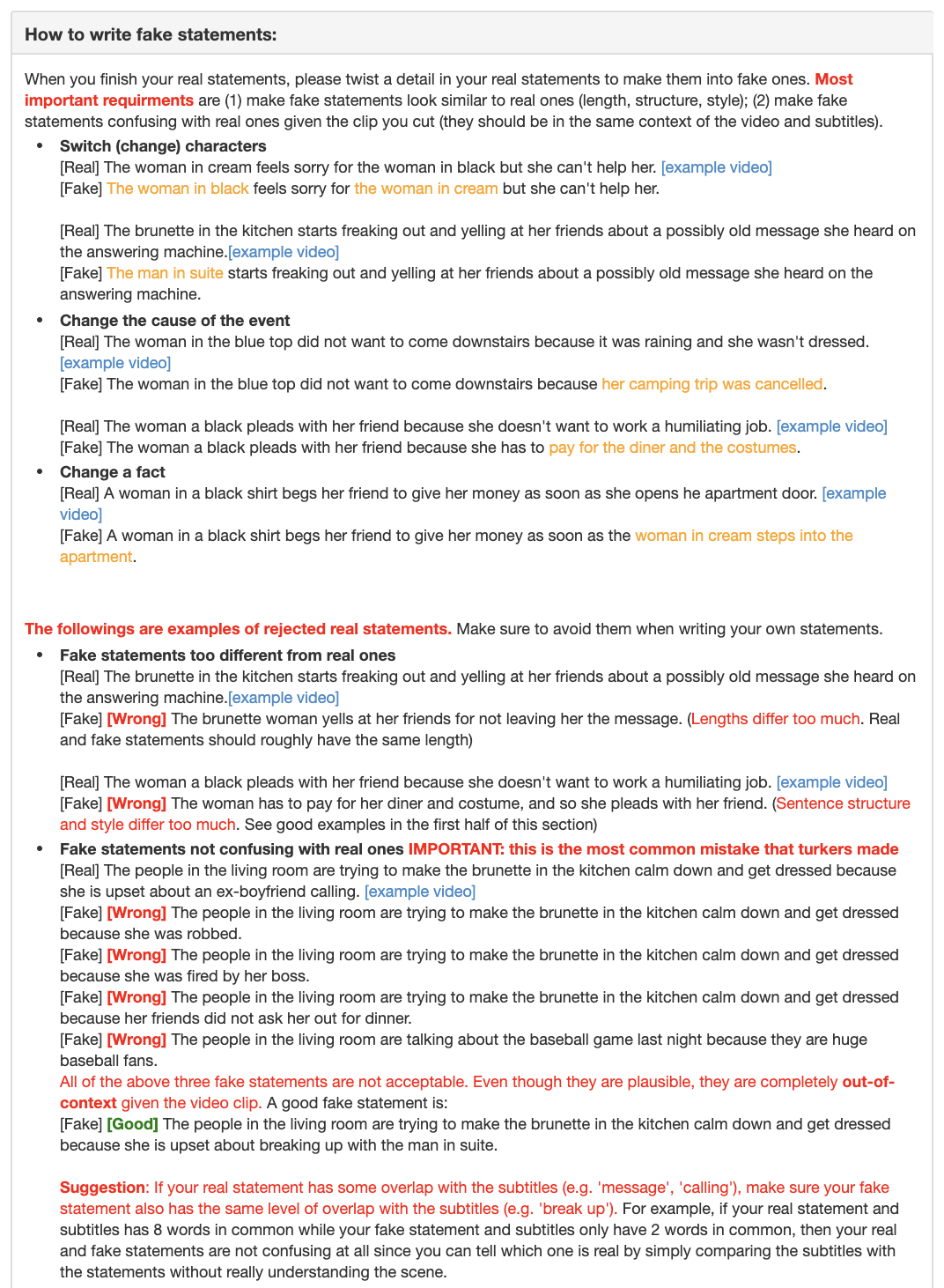}
\end{center}
   \caption{Instructions for writing fake statements.}
\label{fig:ins3}
\end{figure*}

\begin{figure*}[h]
\begin{center}
\includegraphics[width=0.9\linewidth]{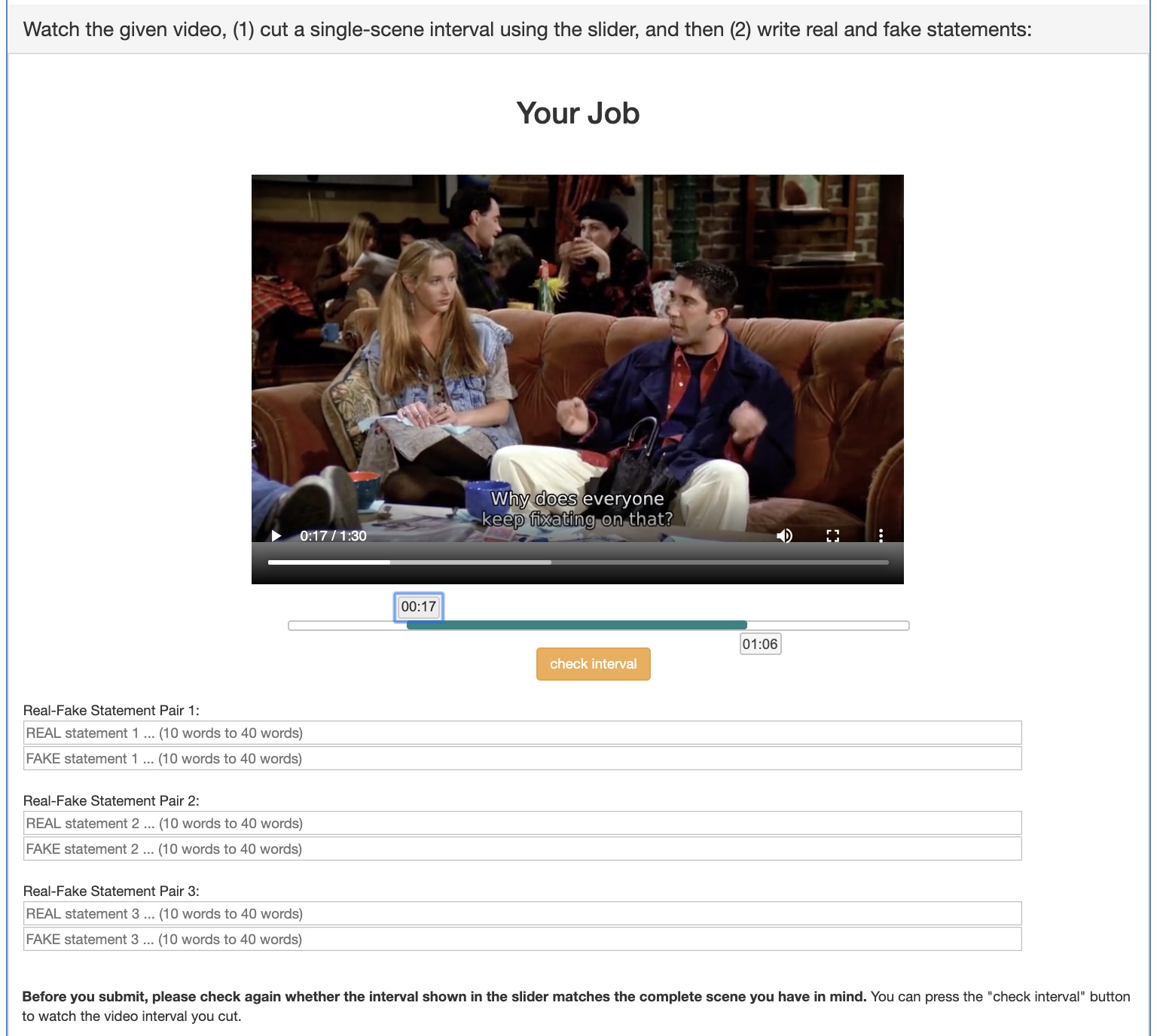}
\end{center}
   \caption{User interface for workers to cut an interval from the video and write three pairs of real and fake statements.}
\label{fig:ins4}
\end{figure*}

Figure~\ref{fig:ins1} through \ref{fig:ins4} show the detailed instructions and user interface for human annotators.

\section{More Examples}
\begin{figure*}[h]
\begin{center}
\includegraphics[width=\linewidth]{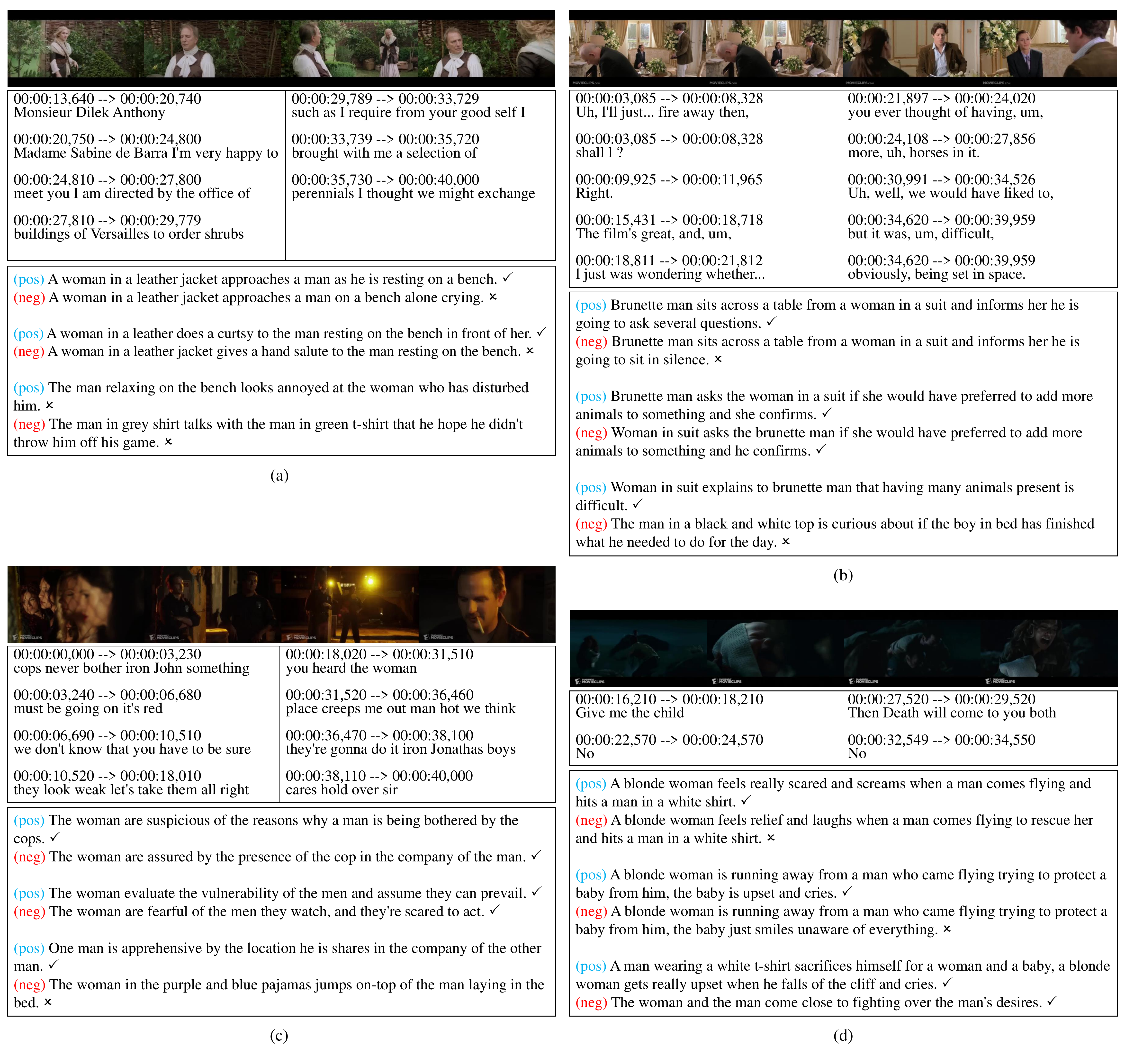}
\end{center}
   \caption{Examples on movie clips. The pos/neg at the beginning of each statement indicates its ground truth. The {\cmark} or {\xmark} at the end of each statement indicates the system's prediction. {\cmark} means the system judges the statement as positive, and {\xmark} means negative.}
\label{fig:movie}
\end{figure*}

\begin{figure*}[h]
\begin{center}
\includegraphics[width=\linewidth]{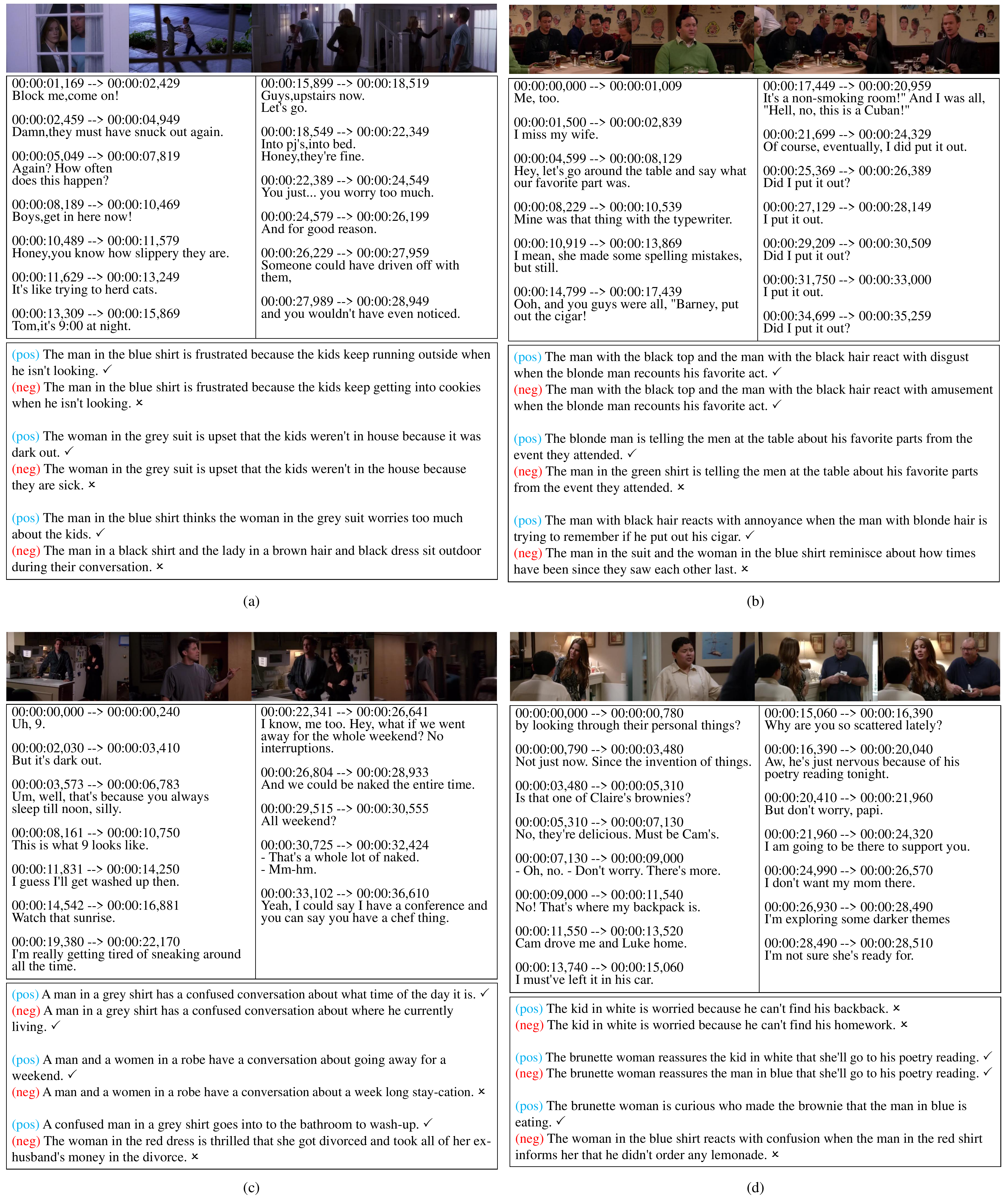}
\end{center}
   \caption{Examples on TV show clips. The pos/neg at the beginning of each statement indicates its ground truth. The {\cmark} or {\xmark} at the end of each statement indicates the system's prediction. {\cmark} means the system judges the statement as positive, and {\xmark} means negative.}
\label{fig:tv}
\end{figure*}

Figure~\ref{fig:movie} and Figure~\ref{fig:tv} show some more examples of predictions from our model on movie and TV show clips. The model used in these examples is Stmt+Subtt+Vis with BERT and Img features.


\end{document}